\documentclass[10pt,twocolumn,letterpaper]{article}

\usepackage{cvpr}
\usepackage{times}
\usepackage{epsfig}
\usepackage{graphicx}
\usepackage{amsmath}
\usepackage{amssymb}
\usepackage{pifont}
\usepackage{subcaption}
\newcommand{\fromTo}[3]{\mathbf{#1}_{\scriptscriptstyle #2\to#3}}

\usepackage[pagebackref=true,breaklinks=true,letterpaper=true,colorlinks,bookmarks=false]{hyperref}

 \cvprfinalcopy 

\def\cvprPaperID{16} 

\ifcvprfinal\fi
\begin{document}

\title{ Multi-Cue Vehicle Detection for Semantic Video
Compression \\ 
in Georegistered Aerial Videos}

\author{Noor Al-Shakarji$^{1,3}$, Filiz Bunyak$^{1}$, Hadi Aliakbarpour$^{1}$, Guna Seetharaman$^{2}$, Kannappan Palaniappan$^{1}$\\
{$^{1}$Electrical Engineering \& Computer Science Department}\\ 
University of Missouri, Columbia, MO, USA 65211 \\
$^{2}$U.S. Naval Research Laboratory, Washington, D.C. \\
$^{3}$University of Technology, Baghdad, Iraq \\
{\tt\small \{nmahyd, bunyak, aliakbarpourh, palaniappank\}@missouri.edu}
}

\maketitle

\begin{abstract}
Detection of moving objects such as vehicles in videos acquired from an airborne camera is very useful for video analytics applications. 
Using fast low power algorithms for onboard moving object detection would also provide region of interest-based semantic information for scene content aware image compression. This would enable more efficient and flexible communication link utilization in low-bandwidth airborne cloud computing networks.
Despite recent advances in both UAV or drone platforms and imaging sensor technologies, vehicle detection from aerial video remains challenging due to small object sizes, platform motion and camera jitter, obscurations, scene complexity and degraded imaging conditions.
This paper proposes an efficient moving vehicle detection pipeline which synergistically fuses both appearance and motion-based detections in a complementary manner using deep learning combined with flux tensor spatio-temporal filtering.
Our proposed multi-cue pipeline is able to detect moving vehicles with high precision and recall, while filtering out false positives such as parked vehicles, through intelligent fusion. 
Experimental results show that incorporating contextual information of moving vehicles enables high semantic compression ratios of over 100:1 with high image fidelity, for better utilization of limited bandwidth air-to-ground network links.


\end{abstract}

\section{Introduction}\label{sec:intro}
Detection of moving vehicles in videos acquired from an airborne camera is very useful for video analytics applications including traffic flow, urban planning, surveillance, law enforcement and disaster response. With the recent advances in sensor technologies and airborne platforms such as unmanned aerial vehicles (UAVs) or drones, there is a growing need for robust video compression, summarization, and automated analysis tools.
The focus of this paper is detection and tracking of moving objects in aerial videos for three types of tasks: (1) {\em video compression} to reduce air-to-ground (UAV/drone to base station) and air-to-air (between drones) communication needs during real-time flight operations; (2) {\em video summarization} to enable efficient inspection of static and dynamic scene content; and (3) {\em semantic video analytics} to derive scene, event, and behavior related actionable knowledge from rich but unstructured video data mining.
 
%
\begin{figure*}[t!]
\begin{center}
   \includegraphics[width=1\linewidth]{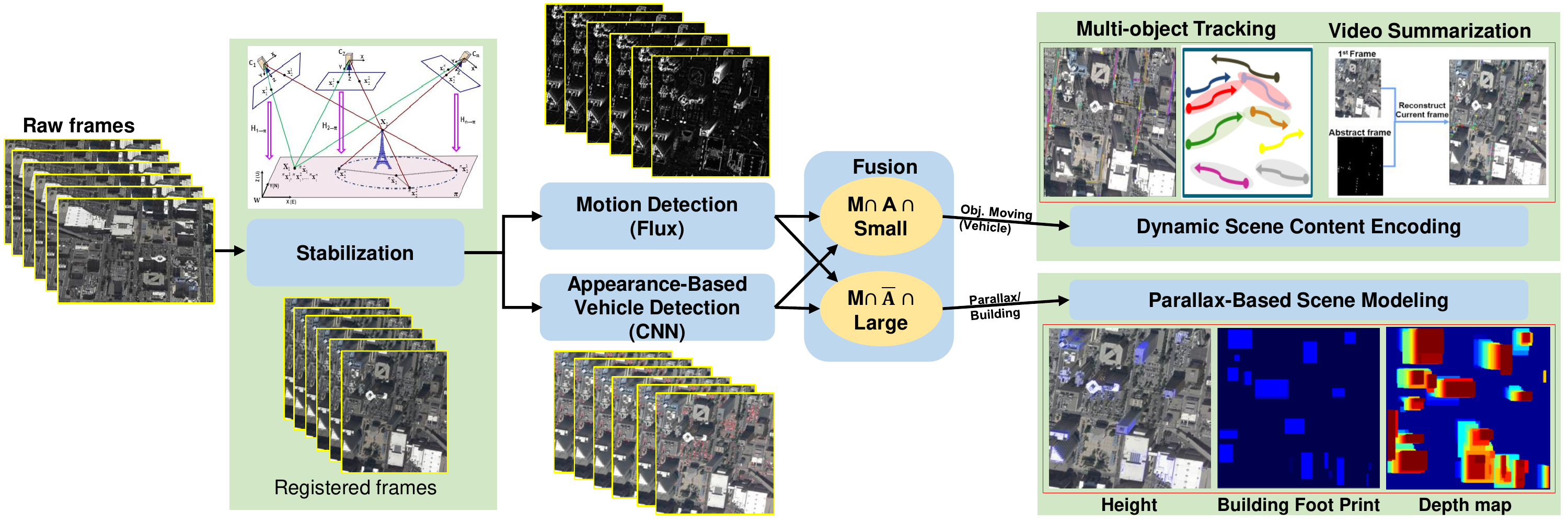}
\end{center}
\vspace{-0.3cm}
   \caption{Multi-cue moving vehicle detection pipeline using motion, appearance and shape information from detections at different stages. In the first stage the aerial video is georegistered and stabilized using \cite{aliakbarpour2017parallax}, in the second stage motion-based flux detection is fused with appearance-based YOLO detections.}
\label{fig:pipeline}
\vspace{-0.6cm}
\end{figure*}
Object detection is at the core of these video analytics tasks.
Advances in deep learning methods, GPU technologies, and training data collected for recent AI challenges~\cite{naphade2017nvidia, zhu2018visdrone, lyu2017ua, lam2018xview} have led to significant performance improvements in object detection accuracy and time efficiency.
Researchers have used motion-based~\cite{farmer2004robust,gautama2002phase,wang2014static} or appearance-based approaches~\cite{basharat2014real,chavez2016multiple} to address the challenges of object detection. Others have combined motion and appearance-based approaches for more robust performance~\cite{siam2017modnet, heo2017appearance,shafiee2017fast}. 
 Despite the improvements, particularly on ground-based video analysis, moving object detection remains a challenging task in wide area motion imagery (WAMI) collected by drones. These videos are characterized by large camera motion, low frame rate, small object sizes, oblique viewing angles, motion blur, parallax effects, shadow and illumination variations, background clutter, partial or full occlusions from buildings, vegetation or other structures, and appearance differences due to weather, environment and seasonal variations. 

This paper proposes a robust moving vehicle detection pipeline for wide area aerial surveillance videos by combining complementary appearance and motion information.
Appearance-based detections are obtained using YOLO (You Only Look Once) ~\cite{redmon2018yolov3} deep learning based object detection system trained with vehicle image patches from aerial imagery. Motion detection is performed using a robust 3D (2D + time) tensor-based approach extending ~\cite{palaniappan2007moving}.
%
There are different sources of motion in an airborne vehicle tracking scenario including: (i) motion of the drone platform itself, (ii) motion of objects (e.g. vehicles and people) in the scene, and (iii) motion induced by parallax due to buildings and other tall structures in the scene.
The platform motion can be eliminated by applying an efficient video registration technique to stabilize the video frames. This step is then followed by a motion detection algorithm to identify moving objects for the purpose of tracking them.
There are several approaches in the literature for motion detection and tracking. However many existing approaches result in enormous amount of false positive detections, due to the spurious motions caused by projection of different views (images acquired from different positions and viewing angles) onto the dominant ground plane (the parallax phenomenon). Classification of real moving objects from parallax induced motions is a very challenging task in WAMI airborne video analysis. In this paper, we introduce a novel pipeline, shown in  Figure~\ref{fig:pipeline}, to identify true moving objects despite spurious detections by fusing deep appearance-based object detection with spatio-temporal tensor-based motion detection. 

Experiments on
Albuquerque urban aerial video dataset (ABQ)~\cite{transSKY} show promising results for detection of not only moving vehicles but also other scene building structures. The paper is organized as follows. Section~\ref{sec:method} describes the details of the proposed pipeline, including main modules for video stabilization, appearance and motion-based detections, and fusion. Section~\ref{sec:exper} presents the experimental results, evaluation methods, and discussion 
of semnatic compression followed by conclusions. 

\section{Multi-Cue Moving Vehicle Detection}
\label{sec:method}
The proposed moving vehicle detection system, for airborne WAMI, consists of four main modules: (1) video stabilization, (2) tensor-based motion detection, (3) appearance-based vehicle detection, and (4) decision fusion. These modules combine computer vision methods (stabilization and motion detection) with machine learning approaches (deep learning for appearance-based detection) and rely on complementary appearance and motion information. Beyond moving vehicle detection, which is the main focus of this paper, the proposed hybrid and multi-cue system also helps in detection of other scene structures such as high-rise buildings that is useful in scene understanding.
\subsection{Video Stabilization Using Georegistration}
Stabilization of sequential video frames is the primary step in many  moving object detection pipelines. Homography is a common method to perform the inter-frame registration and jitter removal. It is often done by estimating a frame-to-frame (piece-wise) perspective transformation (homography) which maps points of an observed dominant plane in the scene from one image's retinal plane to another. Although, the estimation-based methods for obtaining homography transformations may work well for general cases, it becomes very challenging in persistent airborne video (i.e. WAMI) and urban scenery~\cite{Pal:DVSNbook-2011}. This is due to existence of high 3D buildings combined with diversity of the viewing angles causing high level of parallax motion~\cite{Medioni2007}. 
The conventional frame-to-frame homography estimation methods in a long run  are not robust enough and often fail to smoothly  stabilize  the whole sequence of frames without resulting in fragmentations \cite{Linger2015}. 

In our experiments, assuming to have camera 3D poses  (location and orientation) available, a direct analytical homography model is derived. The camera 3D poses can be obtained through different methods such as SLAM \cite{Schneider16} or efficient Bundle Adjustment  \cite{James2017,agarwal2009}.
%
%
\begin{figure}[h!]
	\centerline{
		\includegraphics[width=1.0\columnwidth]{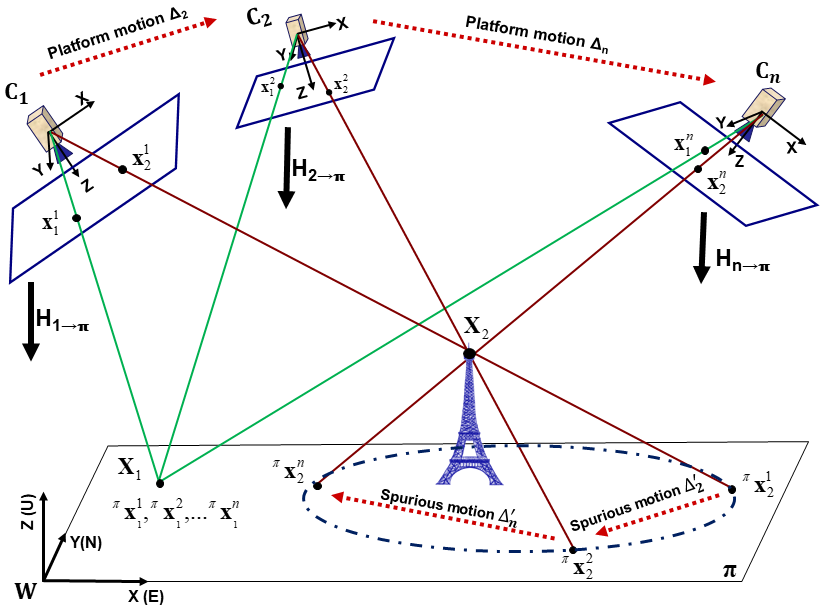}
	}
	\caption{\small{A scene and its dominant ground plane $\pi$ is observed by an airborne camera while hovering over a scene and passing through $n$ way-points. Each image frame is projected using homography onto the scene dominant plane, $\pi$. 
	The homographic transformation of the images of a 3D point like $\mathbf{X}_1$, which lies on plane $\pi$, all converge to an identical 2D point in $\pi$ and are coincident to $\mathbf{X}_1$. Whereas, for an off-plane 3D point such as $\mathbf{X}_2$, its corresponding homography transformations diverge and spread over different locations on $\pi$.
	The diverged points create spurious parallax motions, $\Delta'_2 \dots \Delta'_n$, which can easily be picked up by a motion detection algorithms. The magnitude of such spurious motions are proportional to the height of the 3D structure as well as the platform motion $(\Delta_2 \dots \Delta_n)$.}
	}
	\label{fig:regis_hom1}
	\vspace{-0.35cm}
\end{figure}
Fig. \ref{fig:regis_hom1} shows a world coordinate system $W$  and a dominant ground plane $\pi$ spanning through its $X$ and $Y$ axes. The scene is observed by an airborne camera and images are acquired by the sensor in $n$ way-points along the UAV trajectory. It is equivalent of having $n$ airborne cameras $C_1$, $C_2$ $\dots$ $C_n$. To make the notations succinct, we will omit the camera indices from now on unless otherwise stated.
The image homogeneous coordinate of a 3D point $\mathbf{X}=[ x ~~ y ~~ z  ]^{\intercal}$ 
from the world reference system $W$ projected on the image plane of camera $C$ is obtained as
$\tilde{\mathbf{x}} = \mathbf{K} (\mathbf{R} \mathbf{X} + \mathbf{t})$,
where $\mathbf{K}$ is the calibration matrix (intrinsics), $\mathbf{R}$ and $\mathbf{t}$ are respectively the rotation matrix and translation vector from  $W$ to $C$. For a 3D point $\mathbf{X}$ lying on $\pi$, its $Z$ component is zero, resulting to

\begin{equation}
\tilde{\mathbf{x}} 
=
\mathbf{K} 
\begin{bmatrix}
\mathbf{r}_1 & \mathbf{r}_2 & \mathbf{t}
\end{bmatrix} 
{}^{{\pi}}\tilde{\mathbf{x}}
\label{eq:Pi2C}
\end{equation}

where $\mathbf{r}_1$, $\mathbf{r}_2$ and $\mathbf{r}_3$ are respectively the first, second and third columns of $\mathbf{R}$, and  $^{{\pi}}\tilde{\mathbf{x}}= [ x ~~ y ~~ 1 ]^{\intercal}$ represents the 2D homogeneous coordinates  of the 3D point $\mathbf{X}$ on $\pi$ \cite{Hartley2003}. One can consider the term $\mathbf{K} [ \mathbf{r}_1 ~~ \mathbf{r}_2 ~~ \mathbf{t} ]$ as a $3\times3$ homography transformation matrix which maps any 2D point from $\pi$ onto the camera image plane as:
$\tilde{\mathbf{x}}  =
\fromTo{H}{\pi}{c}
{}^{{\pi}}\tilde{\mathbf{x}}
$.
Likewise, a homogeneous image point $\tilde{\mathbf{x}}$ can be mapped on $\pi$ as
${}^{{\pi}}\tilde{\mathbf{x}}  =
\fromTo{H}{c}{\pi}
\,\tilde{\mathbf{x}}
$, 
where $\fromTo{H}{c}{\pi}$ is the inverse of $\fromTo{H}{\pi}{c}$ and is equal to
\begin{equation}
\fromTo{H}{c}{\pi}
=
\begin{bmatrix}
\mathbf{r}_1 & \mathbf{r}_2 & \mathbf{t}
\end{bmatrix}^{-1}
\mathbf{K}^{-1}. 
\label{eq:Hc2p_initial1}
\end{equation}
Assuming $\mathbf{T}=\begin{bmatrix}
\mathbf{r}_1 & \mathbf{r}_2 & \mathbf{t}
\end{bmatrix}$, $f$ as the focal length in pixel, and $(u,v)$ as the camera image principal point, \eqref{eq:Hc2p_initial1} after simplification can be expressed as:
\begin{equation}
\label{eq:Hc2p_withOmega}
\fromTo{H}{c}{\pi} = \dfrac{1}{\lambda}
\begin{bmatrix} 
~~~m_{11}	&	-m_{21}	&	~~~~ \left[ -m_{11} ~~~~~ m_{21} ~~~~~ m_{31} \right] \mathbf{v}  \\
-m_{12}	&	~~m_{22}	&	~~~~ \left[ ~~~m_{12} ~ -m_{22} ~ -m_{32} \right] \mathbf{v}  \\
~~~~r_{13}	&	~~~r_{23}	&	~~ -\mathbf{r}_3^{\intercal} \mathbf{v}  
\end{bmatrix}
\end{equation}
where $\mathbf{v}=\big[ u ~~ v ~~ f \big]^{\intercal}$ and $\lambda$ is a scalar defined as $
\lambda = f\mathbf{r}_3^{\intercal}\mathbf{t}$,
and $m_{ij}$ is the $minor(i,j)$ of matrix $\mathbf{T}$.
Note that $\lambda$ in \eqref{eq:Hc2p_withOmega} can be omitted as a homography matrix is up-to-scale.

The introduced mathematical model for image stabilization works well for stabilization of parts of the image which lie on the ground dominant plane (on-the-plane). However for off-the-plane points (any non-flat objects such as buildings, cars etc.), their homographic projections introduce significant spurious motions, which can be very distractive for motion detection algorithms.
For example, in Fig. \ref{fig:regis_hom1}, consider $\mathbf{X}_2$ as a 3D point which is off-the-plane. It is imaged as $\mathbf{x}_2^1$, $\mathbf{x}_2^2$ and $\mathbf{x}_2^n$ on the image planes of cameras $C_1$, $C_2$ and $C_n$. Mapping them on $\pi$ using homography transformations will result  ${}^{{\pi}}\mathbf{x}_2^1$, ${}^{{\pi}}\mathbf{x}_2^2$ and ${}^{{\pi}}\mathbf{x}_2^n$.  As illustrated in Fig. \ref{fig:regis_hom1}, these mapped points are all spread out on $\pi$. The magnitude of divergence and the displacement between them is a function of the platform motion,  $\Delta_2 \dots \Delta_n$, and the height of the object (e.g. tall building).
The spurious motions, $\Delta'_2 \dots \Delta'_n$, created from this phenomenon (\textit{Parallax}), are extremely likely to be picked up by motion detection algorithms. 
In our pipeline, this type of spurious motions (parallax induced) are filtered out by using building masks obtained from a 3D model.

\subsection{Tensor-based Motion Detection}
This section describes the tensor-based motion detection module used in the proposed multi-cue pipeline.
Structure tensors for images and video are a matrix representation of partial derivative
information~\cite{palaniappan2007moving}. They allow both orientation estimation and image structure
analysis with applications in image processing and computer
vision. 2D structure tensors have been widely used in edge/corner
detection and texture analysis, and 3D structure tensors have been used
in low-level motion estimation and segmentation \cite{Nath-AVC2005,Nagel:ECCV-1998}.

The 3D structure tensor matrix $\mathbf{J(x)}$ for the spatiotemporal
volume centered at $\mathbf{x}$ can be written in matrix
form, without the positional terms shown, for clarity, as Eq. \ref{2-05}. 
\begin{equation}
\mathbf{J}=\left[\begin{array}{lcccr}
{\int_{\mathbf{\Omega}}}\,\frac{\partial\mathbf{I}}{\partial x}\frac{\partial\mathbf{I}}{\partial x}d\mathbf{y} &  & {\int_{\mathbf{\Omega}}}\,\frac{\partial\mathbf{I}}{\partial x}\frac{\partial\mathbf{I}}{\partial y}d\mathbf{y} &  & {\int_{\mathbf{\Omega}}}\,\frac{\partial\mathbf{I}}{\partial x}\frac{\partial\mathbf{I}}{\partial t}d\mathbf{y}\\
\\
{\int_{\mathbf{\Omega}}}\,\frac{\partial\mathbf{I}}{\partial y}\frac{\partial\mathbf{I}}{\partial x}d\mathbf{y} &  & {\int_{\mathbf{\Omega}}}\,\frac{\partial\mathbf{I}}{\partial y}\frac{\partial\mathbf{I}}{\partial y}d\mathbf{y} &  & {\int_{\mathbf{\Omega}}}\,\frac{\partial\mathbf{I}}{\partial y}\frac{\partial\mathbf{I}}{\partial t}d\mathbf{y}\\
\\
{\int_{\mathbf{\Omega}}}\,\frac{\partial\mathbf{I}}{\partial t}\frac{\partial\mathbf{I}}{\partial x}d\mathbf{y} &  & {\int_{\mathbf{\Omega}}}\,\frac{\partial\mathbf{I}}{\partial t}\frac{\partial\mathbf{I}}{\partial y}d\mathbf{y} &  & {\int_{\mathbf{\Omega}}}\,\frac{\partial\mathbf{I}}{\partial t}\frac{\partial\mathbf{I}}{\partial t}d\mathbf{y}
\end{array}\right]\label{2-05}
\end{equation}
The elements of $\mathbf{J}$ (Eq. \ref{2-05}) incorporate information relating to local, spatial, or temporal gradients. The trace of the structure tensor, $~\mathbf{trace(J)}=\int_{\mathbf{\Omega}}||\nabla I||^{2}d\mathbf{y}$
incorporates total gradient change information in space and time corresponding to \textit{both} moving and non-moving edges of the image sequence, but fails to capture the nature of these gradient changes (i.e. spatial only versus temporal).

The flux tensor \cite{bunyak2007geodesic, bunyak2007flux}, characterizes temporal variations in the optical flow field within a local 3D spatiotemporal volume, and is our extension to 3D structure tensors designed to detect only the moving structures without expensive eigenvalue decompositions. 
In the proposed pipeline, in order to prevent information loss due to isoluminance, we define the \textit{color flux tensor}, $\mathbf{J_{FC}(x)}$, as an extension to the regular flux tensor computed as follows:
\begin{equation}
\scriptsize{
\left[
\begin{array}{ccc}
\displaystyle \sum_{\Omega}\sum_{\small{I=R,G,B}}{\left(I_{xt}\right)^2} 
& 
\displaystyle \sum_{\Omega}\sum_{I=R,G,B}{\left(I_{xt}I_{yt}\right)} 
&
\displaystyle \sum_{\Omega}\sum_{I=R,G,B}{\left(I_{xt}I_{tt}\right)} 
\\
\\
\displaystyle \sum_{\Omega}\sum_{I=R,G,B}{\left(I_{yt}I_{xt}\right)} 
& 
\displaystyle \sum_{\Omega}\sum_{I=R,G,B}{\left(I_{yt}\right)^2} 
&
\displaystyle \sum_{\Omega}\sum_{I=R,G,B}{\left(I_{yt}I_{tt}\right)} 
\\
\\
\displaystyle \sum_{\Omega}\sum_{I=R,G,B}{\left(I_{tt}I_{xt}\right)} 
& 
\displaystyle \sum_{\Omega}\sum_{I=R,G,B}{\left(I_{tt}I_{yt}\right)} 
&
\displaystyle \sum_{\Omega}\sum_{I=R,G,B}{\left(I_{tt}\right)^2} 
\end{array}															
\right]
}
\label{eq:color_flux}
\end{equation}
where the following partial derivative notation is used: 
\begin{align}
\begin{tabular}{r r r}
$I_{x}=\frac{\partial I}{\partial x},$ &
$I_{y}=\frac{\partial I}{\partial y},$ & 
$I_{t}=\frac{\partial I}{\partial t},$ 
\\
$I_{xt}=\frac{\partial^2 I}{\partial x \partial t},$ & 
$I_{yt}=\frac{\partial^2 I}{\partial y \partial t},$  & 
$I_{tt}=\frac{\partial^2 I}{\partial t \partial t}$
\end{tabular}
\end{align} 
The elements of the flux tensor (Eq. \ref{eq:color_flux}) incorporate
information about temporal color gradient changes which leads to efficient
discrimination between stationary and moving image features. The
 trace of the flux tensor matrix, 
\begin{equation}
\mathbf{trace(J_{FC})}=\int_{\mathbf{\Omega}}||\frac{\partial}{\partial t}\nabla\mathbf{I}||^{2}d\mathbf{y}
\label{eq-traceJF}
\end{equation}
can be directly used to classify moving and non-moving regions
without expensive eigenvalue decompositions. 

Both tensor formulations use spatio-temporal consistency efficiently, thus produce less noisy and more spatially coherent edge and motion evidence \cite{Nath-AVC2005}. 
Use of tensor-based edge and motion estimation also allows natural extension to color image processing by taking into account vector nature of color data. Extending differential-based operations to color images is
hindered by the multi-channel nature of color images. The derivatives in different channels can point in opposite directions, hence cancellation might occur by simple addition \cite{van2006robust}.
Use of tensor-based representation  prevents these cancellation effects. 

In the proposed system, color flux tensor is used to identify motion blobs. Since this module is applied after video stabilization module which compensates for camera motion, detected motion blobs predominantly correspond to moving vehicles or parallax caused by high-rise buildings. Both of these structures are of interest for video analytics. Moving vehicles to summarize dynamic content, parallax to summarize static content (buildings) captured by a video. Unfortunately, while successful in detecting these structures, tensor-based motion detection can not distinguish these structures from each other.

\subsection{Appearance-based Vehicle Detection Using Deep Learning}
Recently, deep learning approaches have revolutionized object detection. Faster R-CNN~\cite{girshick2015fast}, YOLO~\cite{redmon2016you}, and SSD~\cite{liu2016ssd} are some of the state-of-the-art object detection methods. Deep learning-based object detectors can be divided into two main categories:  region proposal based detectors (e.g. Faster R-CNN~\cite{girshick2015fast}, R-CNN~\cite{girshick2016region}), and single shot detectors (e.g. YOLO~\cite{redmon2016you}, and SSD~\cite{liu2016ssd}), which do not require a separate region proposal process, making them more computationally efficient. For instance, instead of region proposals, YOLO divides the input image into a grid of cells. 
Real-time moving object detection requires fast and accurate processing.  YOLOv3 ~\cite{redmon2018yolov3}, an extended version of YOLO, is one of the fastest and most accurate object detections networks. It has 53 convolutional layers trained on Imagenet. Then, 53 more layers are stacked to give the full 106 convolutional layers. YOLOv3 performs detection at three different scales by applying $1 \times 1$ detection kernels on feature maps of three different sizes at three different layers in the network. Detecting at different scales improves detection of small objects compared to the previous versions.

\begin{figure}[t!]
\begin{center}
   \includegraphics[width=1\linewidth]{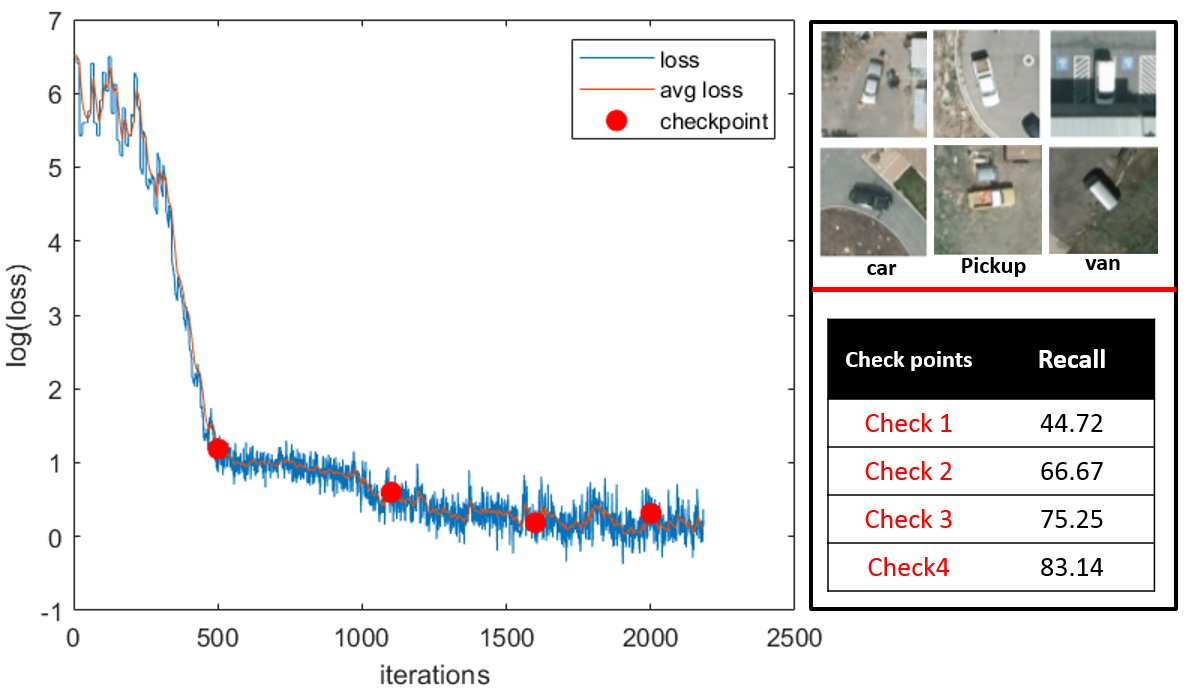}
\end{center}
\vspace{-0.7cm}
   \caption{Loss and average loss for appearance training phase in YOLOv3. The red dots on the curve correspond to recall values listed in the  lower right sub-figure. Sample \textit{vehicle} class subimages for car, pick-up truck and van (3 of 9) VEDAI categories are shown in upper right. }
\label{fig:plots}
\vspace{-0.4cm}
\end{figure}

The annotations for the ABQ dataset used for system test and evaluation in this paper only included moving vehicles. We chose not to train the network with this dataset since the parked vehicles would be considered negative class samples harming the neuron weights during training. Instead we used the Vehicle Detection in Aerial Imagery (VEDAI) dataset ~\cite{razakarivony2016vehicle} for transfer learning by fine tuning the pretrained  YOLOv3 network. VEDAI dataset consists of 1200 satellite imagery collected during Spring 2012, over Utah, USA. The image resolution is $12.5cm \times 12.5cm$
per pixel.  The dataset consists of nine vehicle classes (truck, camping car, tractor, boat, plane, pick-up, car, van and other).  
%
The VEDAI dataset was used to train the appearance-based vehicle detection network. Vehicle class in the proposed system is formed by merging three of the VEDAI subclasses, car, pick-up, and van, into a combined \textit{vehicle} class. Figure \ref{fig:plots} shows loss function for training and some sample image patches for different vehicle types from the VEDAI dataset. 

During training, we set up checkpoints and evaluated the model on the ABQ frames to check the accuracy of training. Figure~\ref{fig:plots} illustrates training progress. The network started to produce reasonable detections starting after 1500 iterations. At each checkpoint we evaluated the accuracy using recall metric (Eq.~\ref{eq:OF5}), which is the ratio of the number of true detected  objects to the total number of ground-truth objects in the dataset.

 Once the YOLOv3 CNN was trained using the VEDAI labeled dataset, then we used our
 ABQ dataset for testing. ABQ WAMI was collected by TransparentSky from an aircraft with on-board GPS and IMU measurements using a circular flight pattern over downtown Albuquerque, NM. 
 %
 Moving vehicles in a subset of 200 cropped frames from the image sequence were manually annotated for testing.
 Parked vehicles were not marked in this labeled ground-truth. 
 ABQ and VEDAI datasets are visually similar in term of object size, scale, and camera viewing angle. Since the images are $2000 \times 2000$ pixels, we divided each frame into 16 non-overlapping $500 \times 500$ patches for higher accuracy in testing the YOLO vehicle detection network. 
 
\subsection{Robust Multi-Cue Moving Vehicle Detection}
\label{sec:fusion}
\begin{figure*}[t!]
\begin{center}
   \includegraphics[width=1\linewidth]{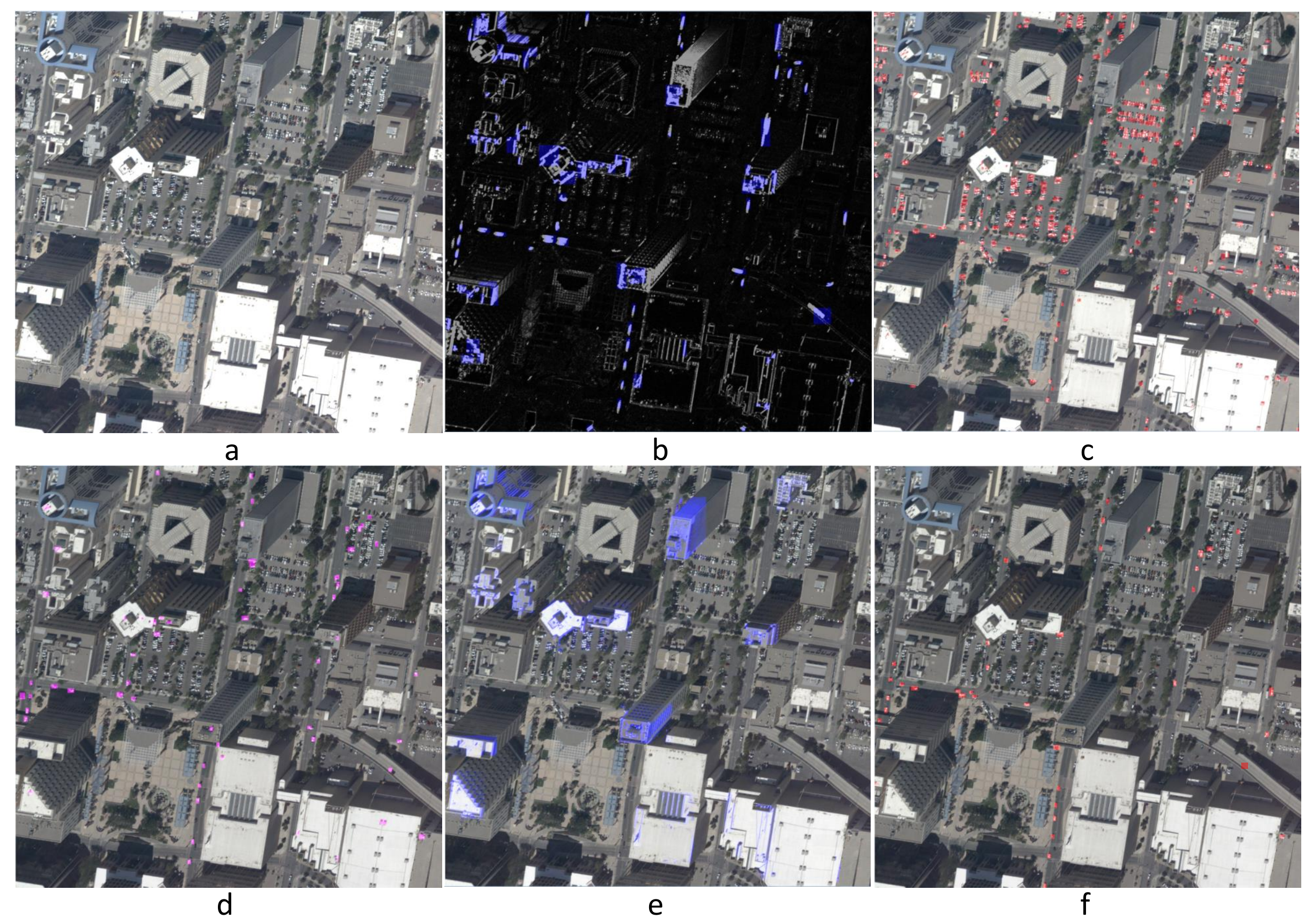}
\end{center}
\vspace{-0.75cm}
   \caption{Intermediate results and the final result after applying the pipeline. a) Raw data, b) Motion mask overlaid on flux tensor motion based detection. c) Appearance mask overlaid on the raw frame, the red overlaid masks represent all predicted vehicles(moved and parked) in the scene. d) Appearance-motion fusion result, some false positive appears on the top of the buildings. e) Buildings mask. f) The final result after filter out false positives on the top of buildings.} 
\label{fig:results}
\vspace{-0.25cm}
\end{figure*}

The goal of the fusion-based multi-cue vehicle decision module is to fuse complementary information from two inherently different approaches to allow semantic classification of motion blobs, filter spurious detections, and boost overall vehicle detection accuracy. Tensor-based motion detection produces spatio-temporally coherent motion detection results robust to illumination changes and soft shadows due to its use of gradient based information. However, since the method relies
on motion, it detects not only moving vehicles but also changes due to motion parallax caused by buildings. Appearance-based detection on the other hand returns only vehicles or other regions with appearances similar to vehicles, whether they are moving or stationary (i.e. parked cars). Stationary cars unnecessarily burden follow-up processes such as communication, tracking, and activity analysis. Unlike ground-based images, where objects with larger support regions have distinct appearance features, WAAS imagery consists of much smaller objects with less distinct features. When trained and tested on these smaller, less distinct objects, false-positives are also most likely compared to their counterparts in ground-based, higher resolution surveillance videos.
Table~\ref{tab:categories} lists the detection categories in the proposed system. 
 Figure \ref{fig:results} illustrates motion-based and appearance-based detection results and fusion outputs for a sample frame. 
 
\begin{figure*}[t!]
\begin{center}
   \includegraphics[width=1\linewidth]{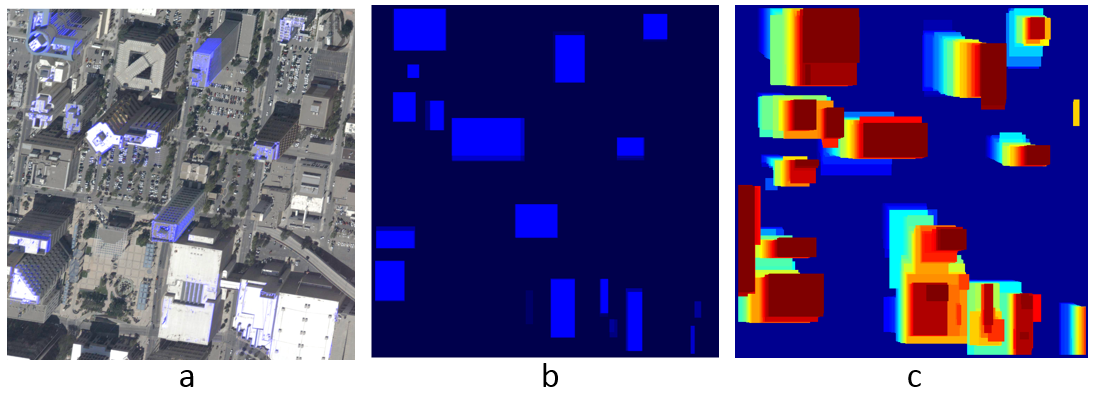}
\end{center}
\vspace{-0.9cm}
   \caption{Building roof-top detection using flux-based motion parallax response. 
   (a) Building parallax response, obtained fusing Flux tensor-based motion and YOLO-based vehicle appearance cues, overlaid on the original frame, 
   (b) building roof-top bounding boxes for a single frame, obtained by post-processing output in (a),  
   (c) per frame building roof-top detections aggregated in time where light blue indicates earlier instances, and red indicates later instances in the image sequence. } 
\label{fig:building}
\end{figure*}

During the fusion process, beside the moving and stationary vehicle category masks, an explicit building category mask is first generated as 
\begin{equation}
    Mask_{Building}=Mask_{Flux}\cap (1-Mask_{YOLO})
\end{equation}
Building mask is then refined by first size based filtering to remove potential false detections, then by morphological operations, connected component labeling, and bounding box fitting (Figure~\ref{fig:building}). While single instance of building roof-top detection is enough to filter-out false vehicle detections. Aggregation of building roof-top detections in time, produces very valuable information regarding 3D scene structure, since spread of the detection instances is directly correlated with building height.
\begin{table}[t!]
    \caption{\small{Fusion procedure for detecting moving vehicles and parallax-based buildings by combining motion (M) and appearance (A) information (see Figure \ref{fig:pipeline}). }}
    \label{tab:categories}
    \centering
    \small{
    \begin{tabular}{|p{0.35in}|p{0.86in}|c|p{1.1in}|}
    \hline
         \bf{Motion (Flux)} & \bf{Appearance (Vehicle CNN)} & \bf{Size} & \bf{Detection Category}\\
         \hline
         \hline
         \text{\sffamily 1} & \text{\sffamily 1} & \text{\sffamily any} & Moving vehicle \\
         \hline
         \text{\sffamily 0} & \text{\sffamily 1} & \text{\sffamily any} & Stationary vehicle or False (obj) detection \\
         \hline
         \text{\sffamily 1} & \text{\sffamily 0} & \text{\sffamily small} & Other moving object or False (motion) detection\\
         \hline
         \text{\sffamily 1} & \text{\sffamily 0} & \text{\sffamily large} & Motion parallax-based buildings\\
         \hline
    \end{tabular}
    }
\end{table}

\section{Experimental results}\label{sec:exper}
The proposed moving vehicle detection system was tested and evaluated on ABQ aerial urban imagery dataset collected using an aircraft with on-board IMU and GPS sensors flying 1.5 km above ground level of downtown Albuquerque, NM. Imaging was done at frame rate of 4Hz and 2.6 km orbit radius. This dataset contains 1071 raw ultra high resolution images ($6400 \times 4400$) with nominal ground resolution of 25cm. Ground-truth for the dataset consists of manually marked bounding boxes and track ids for all the moving vehicles (139 vehicle tracks in total) in $2000 \times 2000$ image patches extracted from 200 consecutive frames.

The results are quantitatively evaluated in terms of detection measures recall, precision (Eq.~\ref{eq:OF5}), and F-measure (Eq.~\ref{eq:OF7}), 
where GT, DT, and TP denote ground-truth, detection, and true prediction objects respectively. 
\begin{equation}
\label{eq:OF5}
Recall= \frac{\# TP}{\# GT}; \quad \quad  Precision= \frac{\# TP}{\# DT}
\end{equation}
\begin{equation}
\label{eq:OF7}
F\-{measure}= 2 \times \frac{Recall \times Precision}{Recall + Precision}
\end{equation}
\begin{table}[t!]
\caption{\small{Precision, recall, and F-measure (in percent) for different stages of the proposed multi-cue moving vehicle detection pipeline. Fusion of motion, vehicle detection and building parallax visual cues yields the highest F-measure.}}
\label{table:RES}
\centering
\scalebox{0.89}{
\begin{tabular}{|l|r|r|r|}
\hline
Detection\_type & Precision  & Recall  & F-measure \\
\hline
\hline
Motion (Flux tensor) & 26.91  & 72.56 & 39.26   \\
Vehicle Appearance (YOLO) & 9.37   & \textbf{83.15} & 16.85   \\
Flux $+$ YOLO  & 53.09  & 71.53 & 60.94   \\
Flux $+$ YOLO $+$ Building & \textbf{69.70}  & 70.53 & \textbf{70.12}\\ 
\hline
\end{tabular}}
\end{table}

\begin{table}[h!]
\caption{\small{Data transfer bandwidth cost measured using different degrees of semantic compression. The columns are the motion detection method, the image types (original image or RGB motion mask), data size in megabytes (MB) and the semantic compression ratio compared to lossless PNG rate.  Mask images have RGB values for motion regions (ROIs) and zero for background pixels.}}
\label{table:commpression}
\centering
\scalebox{0.82}{
\begin{tabular}{|l|l|r|r|}
\hline
\textbf{Motion Detection Cues}  & \textbf{Image Type} &  \textbf{Size}  & \textbf{SCR} \\
\hline
\hline
Original video (Uncompressed) & 200 $\times$ RGBRaw &  2400 & \\
\hline
Original video (PNG, Lossless) & 200 $\times$ RGBPNG &  1070 & 2.2:1\\
\hline
Motion Flux tensor & 1 $\times$ RGBPNG  & 6.0 & \\
      (JPEG, Q=75) & 199 $\times$ RGBMask  &  19.7 & 42:1\\
\hline
Vehicle Appearance YOLO & 1 $\times$ RGBPNG  & 6.0 & \\
    (JPEG, Q=75) & 199 $\times$ RGBMask  &  24.1 & 36:1\\
\hline
Flux + YOLO & 1 $\times$ RGBPNG  & 6.0 &  \\
    (JPEG, Q=75) & 199 $\times$ RGBMask  & 13.0  & 56:1\\
\hline
Flux + YOLO + Building  & 1 $\times$ RGBPNG  & 6.0 & \\ 
  (JPEG, Q=75) & 199 $\times$ RGBMask  & 10.3 &  66:1\\ 
\hline
\end{tabular}
}
\end{table}

Table~\ref{table:RES} 
shows detection performance for different moving vehicle detection approaches.
Motion-only (Flux tensor) detections are shown in Figure~\ref{fig:results}b and Figure~\ref{fig:results}e. Low precision value (26.9\%) for these results are mainly due to false detections caused by motion parallax associated with high-rise buildings. 
Appearance-only using CNN-based detections (YOLOv3) are shown in Figure~\ref{fig:results}c. While, best recall is obtained by this approach, even lower precision (9.4\%) is obtained because of the parked vehicles. %
Combining appearance and motion-based object detections  generates promising results (Figure~\ref{fig:results}d), since parked vehicles get filtered out thanks to the motion mask from the flux tensor. Some false positives still remain due to vehicles parked on building roof tops. Explicit building detection through motion and appearance clues as described in Section~\ref{sec:fusion}, and use of it to further filter vehicles parked on roof tops (Flux + YOLO + Building method) results in the best precision (69.7\%) and F-measure (70.1\%) values.

\begin{figure*}[tbh!]
\begin{center}
   \includegraphics[width=1\linewidth]{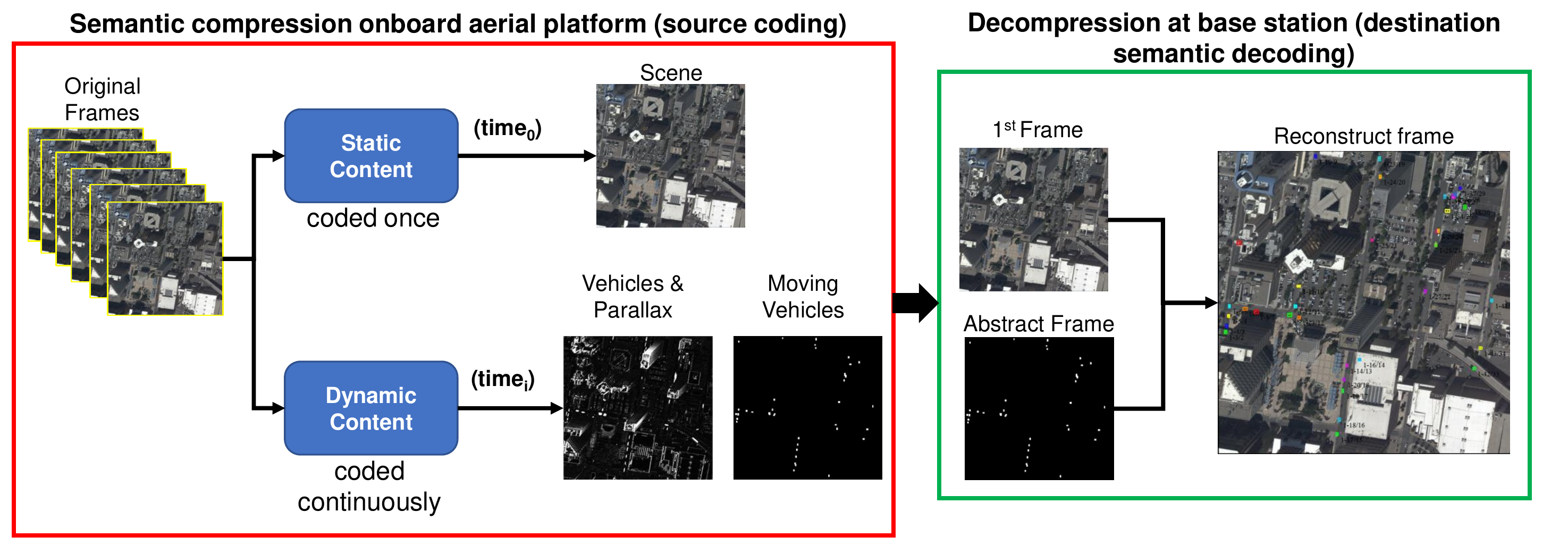}
\end{center}
\vspace{-0.5cm}
   \caption{Semantic compression at the source, onboard an aerial platform, using object detection and embedded processing. Reconstruction of video frames, at the destination, using a base frame combined with semantic information about moving objects (i.e. ROIs) encoded as an \textit{abstract or semantic frame} that is transmitted in a highly compressed format to the base station from the airborne platform.}
\label{fig:summer}
\vspace{-0.5cm}
\end{figure*}

\textbf{Semantic Video Compression} Beside detection measures, we have evaluated the proposed systems in terms of semantic video compression performance. Video compression becomes a very important task during real-time surveillance scenarios where limited communication bandwidth and/or on-board storage greatly restricts air-to-ground and air-to-air communications. 
Efficient handling of video information is important to ensure optimum storage, smoother videos transmission, fast and reliable video processing. 
For considerably reduced communication cost, we propose to transmit the scene information as follows.
The first frame (or another representative frame) from the video is sent from source to destination as a compressed RGB image to represent static content of the scene. Moving vehicle detection is performed on the source platform using one of the proposed methods. Detections are then used to generate mask ROI frames with RGB values for foreground detections, and background pixels set to zero. Following one representative frame, for the remaining frames, only changes are encoded using mask frames (encoding dynamic content of the scene) and transmitted from source to destination.   
Figure~\ref{fig:summer} illustrates encoding and decoding processes at source and destination platforms respectively. The data transfer requirements are shown in Table~\ref{table:commpression} along with the associated semantic compression rate for each method. Combining motion cues enables transmission at higher compression ratios at the same JPEG quality factor. The extracted RGB video frames after georegistration for ABQ are $2000 \times 2000$ for the region of interest and total 2.4GB uncompressed. Mask images have RGB values for motion regions and zero for background pixels. In the proposed semantic compression method, the first frame in the sequence is transmitted as a full color frame using lossless PNG compression (LZ77 dictionary with Huffman coding), to accurately encode scene structures. For the remaining \textit{abstract frames}, ROIs or masks encoding only the changed objects are transmitted using semantic compression ratios of 66:1, greatly reducing the network bandwidth requirements. When the first baseframe does not need to be transmitted, for visualizing moving objects on a map for example, then even higher compression ratios of over 100:1 can be achieved compared to lossless PNG, or 240:1 compared to the uncompressed video stream.

\section{Conclusions}\label{sec:Conclusions}
Object detection is the first step in many advance computer vision applications including multi-object tracking~\cite{al2017robust,al2018multi, al2018robust}, video summarization~\cite{ekin2003automatic, lee2012discovering}, and activity behaviour understanding~\cite{adam2008robust}.
An efficient moving vehicle detection approach from airborne videos was proposed in this paper. We showed that superior performance scores are obtained when a deep learning detection method, YOLO, is fused with a motion based detection algorithm, Flux tensor, in a complementary scheme. 
While all moving and static vehicles are detected by YOLO, fusion of its results with flux tensor as a motion-based detection algorithm allows to considerably eliminate the amount of false alarms. 
Our proposed method produced superior results once applied on different challenging vehicle detection datasets. In addition to vehicle detection and tracking applications, 
the multi-cue approach provides context-based motion blobs for high semantic compression ratios of 100:1, which offers a promising approach to reduce the volume of video data that would need to transmitted between a UAV and a ground station.
In addition to the aforementioned products, more information regarding building structures in the scene, their locations, footprints, and heights can be exploited (as a byproduct) from our pipeline, which could be helpful in situations such as handling occlusions caused by building in airborne videos.
Future work will investigate improving multi-object tracking by incorporating results obtained using the proposed moving vehicle detection system, Flux+YOLO+Building method with our multi-object tracker described in \cite{al2018multi}. This would provide even a higher level of semantic knowledge for achieving greater video compression ratios.

\section{Acknowledgments}
This work was partially supported by awards from U.S. Air Force Research Laboratory FA8750-19-2-0001, 
National Science Foundation CNS-1647084, and CNS-1429294. NAS 
was partially supported by an HCED Government of Iraq doctoral scholarship.
Any opinions, findings, and conclusions or recommendations expressed in this publication are those of the authors and do not necessarily reflect the views of the U.\,S.\ Government or agency thereof.

{\small
\bibliographystyle{ieee_fullname}
\bibliography{egbib}
}

\end{document}